\documentclass{article}
\usepackage{spconf,amsmath,graphicx,hyperref,cite,color,booktabs,tabularx}
\usepackage{tikz}
\usepackage{pifont}

\usepackage{booktabs,tabularx}
\usepackage{caption,bm}
\title{DIRCR: Dual-Inference Rule-Contrastive Reasoning for Solving RAVENs}
\name{Jiachen Zhang$^{\ddag\S}$, Chengtai Li$^{\ddag}$, Jianfeng Ren$^{\ddag\dagger}$, Linlin Shen$^{\S}$, Zheng Lu$^{\ddag \dagger}$, Ruibin Bai$^{\ddag \dagger}$
}
\address{$^{\ddag}$School of Computer Science, University of Nottingham Ningbo China, China \\
$^{\dagger}$Nottingham Ningbo China Beacons of Excellence Research and Innovation Institute, China \\
$^{\S}$School of Artificial Intelligence, Shenzhen University, China
}
\begin{document}
%
\maketitle
\begin{abstract}
Abstract visual reasoning remains challenging as existing methods often prioritize either global context or local row-wise relations, failing to integrate both, and lack intermediate feature constraints, leading to incomplete rule capture and entangled representations. To address these issues, we propose the Dual-Inference Rule-Contrastive Reasoning (DIRCR) model. Its core component, the Dual-Inference Reasoning Module, combines a local path for row-wise analogical reasoning and a global path for holistic inference, integrated via a gated attention mechanism. Additionally, a Rule-Contrastive Learning Module introduces pseudo-labels to construct positive and negative rule samples, applying contrastive learning to enhance feature separability and promote abstract, transferable rule learning. Experimental results on three RAVEN datasets demonstrate that DIRCR significantly enhances reasoning robustness and generalization. Codes are available at \href{https://github.com/csZack-Zhang/DIRCR}{https://github.com/csZack-Zhang/DIRCR}. 
\end{abstract}
\section{Introduction}
\label{sec:intro}
Abstract Visual Reasoning (AVR) is a multidisciplinary field spanning computer science, cognitive science, and artificial intelligence~\cite{malkinski2023review}, focused on deriving abstract concepts from visual data to solve complex problems through pattern recognition and logical inference~\cite{malkinski2023review,he2023hierarchical,he2024data,li2026predictive}. It has served as a benchmark for evaluating human-like generalization in AI systems~\cite{malkinski2023review}, with applications in tasks such as RAVEN-style progressive matrices~\cite{zhang2019raven,hu2021stratified,benny2021scale,li2025darr}, machine number reasoning~\cite{li2025dbcr}, and compositional reasoning~\cite{zerroug2022benchmark,lidsrf}. AVR also supports real-world domains like autonomous driving by enhancing real-time scene interpretation and decision-making~\cite{liang2023visual,grigorescu2020survey,he2025two,10381483}.



AVR models fall into three primary paradigms:
1) Traditional approaches employ relational networks  to model visual relationships~\cite{barrett2018measuring}, or attention-based mechanisms to emphasize salient features~\cite{vaswani2017attention,he2024hierarchical};
2) Neuroscience-inspired models, such as PredRNet, incorporate cognitive principles like prediction and verification for AVR problem-solving~\cite{yang2023neural,li2025dbcr};
3) Graph-based methods use graph neural networks to capture and reason over structural and evolutionary patterns among abstract visual entities~\cite{Wang2023,santoro2017simple}.



However, existing AVR models still face two major challenges:
1) While many methods treat AVR as a direct mapping from all context images to an answer~\cite{zhang2019learning,zheng2019abstract,kim2020few,wang2020abstract}, capturing global patterns but neglecting row-wise\footnote{We focus on row-wise relationships without loss of generality.} analogical relationships, approaches that focus solely on row-wise reasoning~\cite{yang2023neural,li2025dbcr,he2025two} often miss cross-row interactions and global rule consistency. 
2) Supervision is typically confined to the final layer, leaving intermediate features unconstrained~\cite{uselis2025intermediate}. This leads to entangled embeddings that obscure abstract rule distinctions, promoting dependence on superficial features and hindering generalization to unseen data. 


\begin{figure*}[!t] 
    \centering
    \includegraphics[width=\linewidth]{PRPCL.pdf}
    \caption{Block diagram of proposed Dual-Inference Rule-Contrastive Reasoning model: 1) Dual-Inference Reasoning Module (DIRM), integrating a local row-wise analogy path and a global holistic inference path through a gated attention mechanism; and 2) Rule-Contrastive Learning Module (RCLM), regularizing rule-aware features via an auxiliary contrastive loss.}
    \label{fig:DIRCR}
\end{figure*}

To tackle the challenges, we propose a \textbf{D}ual-\textbf{I}nference \textbf{R}ule-\textbf{C}ontrastive \textbf{R}easoning (\textbf{DIRCR}) model for solving RAVEN-style problems. The core of DIRCR is a \textbf{D}ual-\textbf{I}nference \textbf{R}easoning \textbf{M}odule (\textbf{DIRM}), which consists of two complementary reasoning paths:
1) The local path conducts row-wise analogy, predicting the third image from the first two to capture fine-grained intra-row dependencies.
2) The global path holistically infers the missing image using all context images, integrating both intra-row and inter-row relations.
A structure-aware gated attention mechanism further adaptively enhances salient reasoning features and suppresses noise, effectively fusing information from both pathways.

To improve generalization, we introduce the \textbf{R}ule-\textbf{C}ontrastive \textbf{L}earning \textbf{M}odule (\textbf{RCLM}) based on contrastive learning~\cite{li2024regression}. RCLM generates pseudo-labels to form positive samples from the first two rows and the pseudo-positive last row, each representing underlying row-wise rules, while using pseudo-negative candidates as negative samples. These features are projected into a contrastive space, where positives are attracted and negatives repelled. This process regularizes rule-based features via an auxiliary contrastive loss, enhancing discernibility of row-level reasoning patterns and promoting abstract, transferable rule learning.



Our contributions can be summarized as follows:
1) The proposed DIRM integrates global $n$-to-1 reasoning with local row-wise analogy, capturing both long-range consistency and fine-grained relational patterns to more comprehensively exploit multi-level reasoning cues.
2) The RCLM explicitly attracts representations of valid rule combinations while repelling incorrect candidates, enabling learning of abstract and transferable reasoning structures that substantially improve model generalization.
3) DIRCR achieves state-of-the-art performance across three major RAVEN datasets~\cite{zhang2019raven,hu2021stratified,benny2021scale}.



\section{Proposed Method}
\subsection{Overview of DIRCR}

The RAVEN task~\cite{zhang2019raven,hu2021stratified,benny2021scale} is structured as a $3 \times 3$ matrix. Given question images $\{\bm{Q}_{i}\}_{i=1}^8$, the task is to infer the last missing image $\bm{A}^t$ from candidate answers $\{\bm{A}_i\}_{i=1}^8$, so that the three rows of images follow the same underlying rule. 
We propose the DIRCR model (Fig.~\ref{fig:DIRCR}) to solve it. A lightweight ResNet-4b serves as the feature encoder. The DIRCR consists of two core modules. 
1)~The DIRM uses a dual-path design: the global path infers the missing image from all context panels, and the local path performs row-wise analogy to model intra-row relations. Both paths employ convolutional layers for prediction, compare results with targets, and utilize prediction errors to capture latent rules. DIRM further integrates error-guided gated attention to amplify rule-consistent features and suppress noise, effectively merging global and local cues to learn structural and analogical patterns. 
2)~The RCLM employs pseudo-label rule-contrastive learning to enhance representations. It generates pseudo-labels for candidate answers, forming positive samples from the first two rows and the pseudo-positive third row, and negative samples from pseudo-negative third-row candidates. All are projected into a contrastive space, where positive pairs are attracted and negative pairs repelled. This promotes compact clustering of valid rule structures and separation of incorrect ones, enforcing row-level structural consistency.

\subsection{Dual-Inference Reasoning Module}
\noindent\textbf{Local Row-wise Reasoning:} \quad 
Row-wise rule modelling has been widely adopted~\cite{yang2023neural,li2024regression,he2025two}. 
In particular, for each row $(i)$, the prediction network \(\psi_L(\cdot)\) predicts the last column of features using the first two columns of features as:
\begin{align}
\hat{\bm{F}}_{3}^{(i)} = \psi_L(\bm{F}_{1}^{(i)}, \textbf{F}_{2}^{(i)}).
\end{align}
The predictor \(\psi_L(\cdot)\) uses a 2-D convolution, batch normalization, and ReLU activation (ConvNormAct) for stable training and convergence. All rows share the same predictor, ensuring representational consistency, reducing parameters, and promoting learning of general compositional patterns (row-specific features) to improve generalization. 
The predicted target feature is subtracted from the actual target feature to obtain the prediction error: \(\bm{E}^{(i)} = \hat{\bm{F}}_{3}^{(i)} - \bm{F}_{3}^{i}\). 
The residual features are concatenated with the original features along the channel dimension to form the output feature \(\bm{O}_L\). This residual connection preserves original semantics, preventing over-reliance on residuals while enhancing discriminative representation learning~\cite{li2024regression}.

\noindent\textbf{Global $n$-to-1 Reasoning:} \quad 
DIRM incorporates both global $n$-to-1 and local row-wise reasoning. The global predictor \(\psi_{G}(\cdot)\) infers the target feature using all eight context images:
\begin{align}
\hat{\bm{F}}_{t} = \psi_{G}(\bm{F}_{1}, \dots, \bm{F}_{8}),  
\end{align}
and the residual \(\bm{E}_{G} = \hat{\bm{F}}_{t}-\bm{F}_{t}\) is computed and fused with the original features as in the row-wise branch to integrate holistic contextual patterns to form the output feature \(\bm{O}_G\).

\noindent\textbf{Gated Attention Mechanism:} \quad
To enhance interaction between local row-wise and global residuals, we introduce a cross-attention fusion mechanism. The concatenated outputs 
\(\bm{O}_L\) (from local reasoning branch) and \(\bm{O}_G\) (from global reasoning branch) are processed through multi-head cross-attention \(\mathcal{F}_\textsc{CA}(\cdot)\) and self-attention \(\mathcal{F}_\textsc{SA}(\cdot)\) as:  
\begin{align}
\bm{O} = \mathcal{F}_\textsc{SA}(\mathcal{F}_\textsc{CA}(\bm{O}_L,\, \bm{O}_G)+\mathcal{F}_\textsc{CA}(\bm{O}_G,\, \bm{O}_L)).
\end{align}
The three attention layers iteratively fuse local (\(\bm{O}_L\)) and global  (\(\bm{O}_G\)) reasoning features via bidirectional interaction and integrated refinement. This enhances semantic consistency, reasoning robustness, and generalization. 

Then, a structure-aware gating mechanism dynamically modulates features using a linearly projected gating vector \(\bm{g}\). The fused output \(\bm{O}\) is element-wise modulated as:
\begin{align}
\bm{R} = \gamma(\bm{g}) \odot \bm{O},
\end{align}
where \(\gamma\) is the \texttt{GELU} activation and \(\odot\) is the element-wise multiplication. This mechanism enhances rule-consistent features while reducing noise, focusing on valid patterns and improving robustness and generalization in abstract reasoning.

\noindent\textbf{Hierarchical Reasoning:} \quad 
To handle increasingly complex structural rules and their interactions, we stack \(K\) DIRMs, enabling deeper reasoning over compositional relationships: 
\begin{equation}
\bm{R}^{(k)} = \textsc{DIRM}^{(k)}\big(\bm{R}^{(k-1)}\big), 
\qquad k = 1,\dots,K,
\end{equation}
This hierarchical design facilitates progressive learning from simple patterns to higher-order rule combinations, enhancing representational capacity and structural generalization.


\subsection{Rule-Contrastive Learning Module}
After training the DIRM, we can utilize it to generate the likelihood $\bm{p}_{j}$ of each candidate image $\bm{A}_j$ being the correct one so that the last row of three images follows the same rule as the first two rows. Then, we select the candidate with the highest probability as the pseudo-label \(\hat{c} = \operatorname{argmax}_{j} \bm{p}_{j}\). 




After obtaining pseudo-labels, we construct positive and negative sample sets using reasoning features as:
\[
\mathcal{P} = \{\bm{G}_1,\; \bm{G}_2,\; \bm{G}_{3}^{\hat{c}}\},
\qquad
\mathcal{N} = \{\bm{G}_{3}^j \mid j\neq\hat{c}\}. 
\]
where \(\bm{G}_1\) and \(\bm{G}_2\) denote the features of the first and second rows, and \(\bm{G}_{3}^{\hat{c}}\) represents the third-row features with the pseudo-positive candidate, and \(\bm{G}_{3}^j\) denotes the third-row features with a pseudo-negative candidate. These row-wise features encode the latent relations within each triple-image row.

A two-layer perceptron maps row-wise reasoning features into a contrastive space to reduce classification bias and semantic noise. Supervised contrastive loss enhances discriminability by maximizing intra-positive similarity and minimizing positive-negative affinity, pulling positives together and pushing negatives apart per anchor. 
\[
\mathcal{L}_{\mathrm{RCLM}} = -\sum_{\bm{P}\in\mathcal{P}} \log \frac{\displaystyle\sum_{\bm{U}\in \mathcal{P}} \mathrm{sim}(\bm{P},\bm{U})} {\displaystyle\sum_{\bm{U}\in \mathcal{P}\cup \mathcal{N}_i} \mathrm{sim}(\bm{P},\bm{U})}, 
\]
where \(\mathrm{sim}(\cdot,\cdot)\) denotes a similarity function. This restructures intermediate features: rule-consistent positives cluster tightly, while rule-violating negatives are separated, reducing ambiguity from end-layer-only supervision. 

\section{Experimental Results}

\subsection{Experimental Settings}




We conduct comparison experiments on the RAVEN~\cite{zhang2019raven}, I-RAVEN~\cite{hu2021stratified} and RAVEN-FAIR~\cite{benny2021scale} datasets. 
These three datasets have been widely utilized to evaluate the reasoning capabilities. The DIRCR model is compared with ten state-of-the-art methods, including LEN~\cite{zheng2019abstract}, SRAN~\cite{hu2021stratified}, DCNet~\cite{zhuo2022effective}, MRNet~\cite{benny2021scale}, SCL~\cite{wu2020scattering}, Rel-Base~\cite{spratley2020closer}, STSN~\cite{mondal2023learning}, PredRNet~\cite{yang2023neural}, DRNet~\cite{zhao2024learning} and TRIVR~\cite{he2025two}. 
Following the standard evaluation protocol~\cite{zhang2019raven, hu2021stratified, benny2021scale}, the number of training, validation, and test samples in each dataset is 42,000, 14,000, and 14,000, respectively.

Images are resized to \(80 \times 80 \times 1\), with a batch size of 128. Training uses the Adam optimizer (initial learning rate \(10^{-3}\), weight decay \(10^{-5}\)). The model employs $K=3$ DIRMs and a dropout rate of 0.1 in both reasoning and classification layers. Rule-Contrastive Learning (RCL) is applied with a projection dimension of 128, temperature \(\tau = 0.2\), confidence threshold 0.60, and loss weight 0.1.

\subsection{Comparison to State-of-the-Art Methods}


\noindent\textbf{Comparison on RAVEN/I-RAVEN/RAVEN-FAIR Dataset:} \quad 
Tab.~\ref{tab:SOTA} reports the reasoning accuracy (\%) of various models on RAVEN, I-RAVEN, and RAVEN-FAIR. Key observations include:
1)~DIRCR consistently achieves state-of-the-art performance across all datasets, with 98.5\% on RAVEN, 97.8\% on I-RAVEN, and 98.7\% on RAVEN-FAIR (avg: 98.3\%), outperforming strong baselines like STSN~\cite{mondal2023learning} (96.4\%), DRNet~\cite{zhao2024learning} (97.4\%), and TRIVR~\cite{he2025two} (94.8\%). 
2)~The significant performance gains on the bias-minimized datasets RAVEN and RAVEN-FAIR further demonstrate that DIRCR is more robust to superficial visual biases and excels at extracting underlying structural rules for abstract reasoning. 
3)~Despite the strong performance of existing methods, DIRCR achieves substantial improvements over state-of-the-art models, attributable to its dual-inference architecture and rule-contrastive learning mechanism.

\begin{table}[!t]
\captionsetup{width=\columnwidth}
  \caption{Comparison with state-of-the-art models on the RAVEN~\cite{zhang2019raven}, I-RAVEN~\cite{hu2021stratified} and RAVEN-FAIR~\cite{benny2021scale} datasets.}
  \centering
  \setlength{\tabcolsep}{3pt}
  \begin{tabular}{lllll}
    \toprule
    Model & RAVEN & I-RAVEN & RAVEN-F & Avg \\
    \midrule
    LEN~\cite{zheng2019abstract}\textsubscript{NeurIPS'19}       & 72.9 & 41.4 & 51.0 & 55.1 \\
    SRAN~\cite{hu2021stratified}\textsubscript{AAAI'21}       & 54.3 & 72.9 & 60.8 & 62.7 \\
    DCNet~\cite{zhuo2022effective}\textsubscript{ICLR'21}     & 93.6 & 56.1 & 47.2 & 65.6 \\
    MRNet~\cite{benny2021scale}\textsubscript{CVPR'21}        & 96.6 & 88.4 & 83.5 & 89.5 \\
    SCL~\cite{wu2020scattering}\textsubscript{arXiv'23}        & 91.7 & 93.5 & 91.1 & 92.1 \\
    Rel-Base~\cite{spratley2020closer}\textsubscript{ECCV'20} & 91.6 & 90.1 & 95.0 & 92.2 \\
    STSN~\cite{mondal2023learning}\textsubscript{ICLR'23}     & 95.8 & 97.1 & 96.5 & 96.4 \\
    PredRNet~\cite{yang2023neural}\textsubscript{ICML'23}     & 89.7 & 95.4 & 95.7 & 93.6 \\
    DRNet~\cite{zhao2024learning}\textsubscript{AAAI'24}      & 96.9 & 97.6 & 97.8 & 97.4 \\
    TRIVR~\cite{he2025two}\textsubscript{PR'25}           & 93.6 & 95.9 & --   & 94.8 \\
    \midrule
    \textbf{DIRCR}                     & \textbf{98.5} & \textbf{97.8} & \textbf{98.7} & \textbf{98.3} \\
    \bottomrule
  \end{tabular}
  \label{tab:SOTA}
\end{table}

\begin{table}[t] 
  \centering
  \captionsetup{width=\columnwidth}
  \setlength{\tabcolsep}{3pt}
  \caption{Comparisons on RAVEN and I-RAVEN shrunk datasets following the setups in~\cite{he2025two}.} 
  \label{tab:rvp}
  \begin{tabular}{llll}
    \toprule
    \textbf{Models} & 10,500 & 21,000 &42,000\\
    \midrule
    CoPINet~\cite{zhang2019learning}\textsubscript{NeurIPS'19}     & 81.8/- &89.2/-&91.4/50.6\\
    SRAN~\cite{hu2021stratified}\textsubscript{AAAI'21}        & -/40.8 &-/53.4&54.3/72.9 \\
    SCL~\cite{wu2020scattering}\textsubscript{arXiv'23}          & 64.8/68.4 & 75.0/78.7&91.7/93.5\\
    TRIVR~\cite{he2025two}\textsubscript{PR'25}          & 91.5/94.7 & 91.8/94.9&93.6/95.9\\
    \midrule
    \textbf{DIRCR} & \textbf{92.3/95.8} & \textbf{94.1/97.1} & \textbf{98.5/97.8} \\
    \bottomrule
  \end{tabular}
  \label{tab:Shrunk}
\end{table}



\noindent\textbf{Comparison on Shrunk Datasets:} \quad 
DIRCR effectively extracts latent rules from abstract reasoning problems, maintaining strong performance even under limited training data. To evaluate this, we conducted comparisons on shrunk RAVEN/I-RAVEN datasets following~\cite{he2025two}, utilizing only 1/4 (10,500 samples) and 1/2 (21,000 samples) of the original training data, while keeping validation and test sets unchanged. 
Tab.~\ref{tab:Shrunk} shows that DIRCR achieves state-of-the-art performance across shrunk RAVEN and I-RAVEN. It attains 92.3\%, 94.1\%, and 98.5\% on RAVEN at 1/4, 1/2, and full training scales, outperforming TRIVR~\cite{he2025two} and CoPINet~\cite{zhang2019learning}. On I-RAVEN, it reaches 95.8\%, 97.1\%, and 97.8\%, exceeding TRIVR~\cite{he2025two} and SCL~\cite{wu2020scattering}, demonstrating strong generalization. DIRCR also shows better robustness to data bias, with less accuracy drop on debiased I-RAVEN, indicating greater reliance on structural rules. Moreover, it achieves high accuracy (92.3\% on RAVEN, 95.8\% on I-RAVEN) with only 10,500 samples, highlighting high data efficiency. 

\subsection{Ablation Studies of Key Modules and $K$}
We conducted ablation studies on three RAVENs to evaluate the contribution of each module in DIRCR and the effect of the number of DIRM blocks. Tab.~\ref{tab:3} reveals the following key findings:
1)~Using only 2-to-1 or 8-to-1 modules yields the lowest accuracies (97.0\% and 96.5\%, respectively), with 2-to-1 slightly outperforming 8-to-1;
2)~Adding RCLM to the 2-to-1 module improves accuracy to 97.3\%, with notable gains on I-RAVEN and RAVEN-FAIR;
3)~Combining 2-to-1 and 8-to-1 modules integrates local and global information, substantially boosting performance;
4)~Employing all three modules achieves the highest accuracy by capturing both structural patterns and deeper abstract rules;
5)~Increasing DIRM blocks from $K=1$ to $K=3$ improves accuracy from 97.5\% to 98.3\%, suggesting iterative reasoning enhances modeling of complex rules. However, performance drops to 97.6\% at $K=4$, likely due to overfitting.

\begin{table}[!t]
  \caption{Ablation of the two major components and \(K\).  
  }
  \centering
  \setlength{\tabcolsep}{2pt}
  \begin{tabular}{ccc|cccc}
    \toprule
    2-to-1 & 8-to-1 & RCLM & RAVEN & I-RAVEN & RAVEN-F & \textbf{Avg.}\\
    \midrule
    \ding{51}&\tikz[baseline=(char.base)]{\node[inner sep=0pt, outer sep=0pt, text opacity=0.3] (char) {\ding{55}};} & \tikz[baseline=(char.base)]{\node[inner sep=0pt, outer sep=0pt, text opacity=0.3] (char) {\ding{55}};} 
              &96.2 &97.3 &97.6 &97.0 \\
    \tikz[baseline=(char.base)]{\node[inner sep=0pt, outer sep=0pt, text opacity=0.3] (char) {\ding{55}};} &\ding{51} 
             & \tikz[baseline=(char.base)]{\node[inner sep=0pt, outer sep=0pt, text opacity=0.3] (char) {\ding{55}};} &96.0 &96.4 &97.1 &96.5 \\
    \ding{51} 
              & \tikz[baseline=(char.base)]{\node[inner sep=0pt, outer sep=0pt, text opacity=0.3] (char) {\ding{55}};}  &\ding{51} &96.5 &97.7 &97.8&97.3
              \\
    \ding{51} & \ding{51} &\tikz[baseline=(char.base)]{\node[inner sep=0pt, outer sep=0pt, text opacity=0.3] (char) {\ding{55}};} &97.1 &97.5&98.1&97.6 \\
    \ding{51} & \ding{51} &\ding{51} &\textbf{98.5} &\textbf{97.8}&\textbf{98.7}&\textbf{98.3}\\
\midrule \midrule
\multicolumn{3}{c|}{\(K=1\)} &96.9 &97.4 &98.2&97.5  \\ 
\multicolumn{3}{c|}{\(K=2\)} &97.3 &97.5 &98.3&97.7  \\ 
\multicolumn{3}{c|}{\(K=3\)} &\textbf{98.5} &\textbf{97.8}&\textbf{98.7}&\textbf{98.3}\\ 
\multicolumn{3}{c|}{\(K=4\)} &96.7 &97.7 &98.4 &97.6 \\
    \bottomrule
  \end{tabular}
\label{tab:3}
\end{table}


\section{Conclusion}
To advance the comprehension of abstract rules and enhance robust reasoning capabilities, this paper introduces the Dual-Inference Rule-Contrastive Reasoning (DIRCR) model, consisting of two core modules. The DIRM module integrates global structural reasoning with local analogical reasoning, while the RCLM module imposes contrastive constraints by attracting representations of valid rule combinations and repelling incorrect candidates. Extensive experiments on three RAVENs show that DIRCR consistently and significantly outperforms state-of-the-art methods across diverse settings. 

\clearpage
\newpage

\small
\bibliographystyle{IEEEbib}
\bibliography{strings,refs}

\end{document}